\documentclass{article}
\usepackage{spconf,graphicx,hyperref}
\usepackage{graphicx}
\usepackage{subcaption} 
\usepackage{cite}
\usepackage{amsmath,amssymb,amsfonts}
\usepackage{algorithmic}

\usepackage{textcomp}
\usepackage{xcolor}

\title{Active Inference-Driven World Modeling for Adaptive UAV Swarm Trajectory Design}
\name{ Kaleem Arshid $^{1,2}$, Ali Krayani $^{1}$, Lucio Marcenaro $^{1}$, David Martin Gomez $^{2}$, Carlo Regazzoni $^{1}$ \thanks{This work was partially supported by the European Union under the Italian National Recovery and Resilience Plan (PNRR) of NextGenerationEU partnership on "Telecommunications of the Future" (PE00000001 - program "RESTART"), CUP E63C22002040007 - D.D. n.1549 of 11/10/2022, and in part by the Ministry of University and Research (MUR), National Recovery and Resilience Plan (NRRP), Mission 4, Component 2, Investment 1.5, project ”RAISE - Robotics and Al for Socio-economic Empowerment” (ECS00000035).}} 

\address{$^{1}$University of Genoa, Italy, $^{2}$ Carlos III University Madrid, Spain}
\begin{document}
%
\maketitle

\begin{abstract}
This paper proposes an Active Inference–based framework for autonomous trajectory design in UAV swarms. The method integrates probabilistic reasoning and self-learning to enable distributed mission allocation, route ordering, and motion planning. Expert trajectories generated using a Genetic Algorithm with Repulsion Forces (GA–RF) are employed to train a hierarchical World Model capturing swarm behavior across mission, route, and motion levels. During online operation, UAVs infer actions by minimizing divergence between current beliefs and model-predicted states, enabling adaptive responses to dynamic environments. Simulation results show faster convergence, higher stability, and safer navigation than Q-Learning, demonstrating the scalability and cognitive grounding of the proposed framework for intelligent UAV swarm control.
\end{abstract}


\section{Introduction}
Unmanned aerial vehicle (UAV) swarms have gained significant attention for their potential in distributed autonomy, scalability, and cooperative decision-making \cite{Ahmad2025, ghazali2021systematic}. Their applications extend from surveillance and mapping to complex multi-agent missions such as coordinated inspection, transportation, and communication support. The key challenge is to enable UAVs to collaboratively plan energy-efficient, collision-free, and adaptive trajectories while preserving global mission objectives \cite{javed2024state, arshid2025toward}.  

Classical optimization methods such as Mixed Integer Programming and graph-based formulations \cite{liu2021car, chai2025trajectory} offer deterministic solutions but rely on full system knowledge, limiting their use under uncertainty. Metaheuristic approaches (e.g., Genetic Algorithms, Particle Swarm, Ant Colony Optimization) \cite{wang2023improved, gad2022particle, manullang2023optimum} are flexible yet require repeated recomputation for new missions. Data-driven techniques like Deep or Multi-Agent Reinforcement Learning \cite{shi2024deep, li2023computation} achieve autonomy through experience but demand extensive training and exhibit weak generalization to unseen scenarios.  

Recent advances in probabilistic generative modeling and cognitive inference frameworks aim to unify perception, reasoning, and control \cite{pezzulo2024active, nozari2022incremental}. Among them, \textit{Active Inference} provides a principled Bayesian mechanism for adaptive behavior by continuously minimizing the divergence between predicted and observed states.  

Building on this paradigm, this work introduces an Active Inference–based framework for UAV swarm trajectory design. The approach combines probabilistic reasoning with hierarchical symbolic decision-making to enable self-consistent, multi-level adaptation in real time. It unifies model-based structure and learning-based flexibility, allowing UAVs to reason jointly about mission allocation, route ordering, and motion generation.  

The main contributions are summarized as follows:  
\textit{(i)} a \textbf{World Model} representing swarm behavior across mission, route, and motion levels;  
\textit{(ii)} a \textbf{Probabilistic Decision-Making Mechanism} based on Active Inference for adaptive planning; and 
\textit{(iii)} a \textbf{Filter-Assisted Control Strategy} using Extended Kalman Filtering (EKF) for smooth and safe navigation. This unified probabilistic–symbolic framework provides interpretable, adaptive, and scalable trajectory design for complex, uncertain environments.
%
%
\section{System and Problem Setup}
%
\vspace{-3mm}
We consider a swarm of $Q$ UAVs, $U=\{u_1,\dots ,u_Q\}$, deployed to cooperatively visit a set of $N$ target locations $C=\{c_1,\dots ,c_N\}$ within a bounded area. Each UAV $u_q$ starts and ends at a common depot $L_0=[x_0,y_0,z_0]$ and is described by its position $\mathbf{x}_q(t)=[x_q,y_q,z_q]$, velocity $v_q(t)$, and heading change $\Delta\phi_q(t)$. UAVs maintain altitude limits and a minimum inter-UAV distance $d_{\min}$ to ensure safe operation.
The swarm must ensure that each target $c_i\!\in\!C$ is visited once by exactly one UAV while minimizing overall mission cost (distance, time, or energy). The decision process is organized hierarchically:
\\
\noindent
\textbf{• High level – Mission allocation:}
partition $C$ into disjoint subsets $\{C_1,\dots ,C_Q\}$, one per UAV, balancing workload and cost.\\
\textbf{• Medium level – Route sequencing:}
for each UAV $u_q$, determine the visiting order $\pi_q=[c_{q,1},\dots ,c_{q,|C_q|}]$, where $|C_q|$ is the total number of cities in the subset $C_q$.\\
\textbf{• Low level – Trajectory generation:}
compute dynamically feasible, collision-free paths between consecutive towns.

Let $d_{ij}$ denote the Euclidean distance between $c_i$ and $c_j$.
The overall coordination problem is formulated as a \textit{multi-traveling-salesman problem (MTSP)}:
\begin{equation}
\small
    \min_{X^q_{ij}} J=\sum_{q=1}^{Q}\sum_{i,j=1}^{N} d_{ij}X^q_{ij},
    \label{eq:mtsp}
\end{equation}
subject to standard visit-once, flow-conservation, and subtour-elimination constraints, and to UAV dynamics
\begin{equation}
\small
    v_q(t)\!\le\!v_q^{\max},\;
    |\Delta\phi_q(t)|\!\le\!\phi_q^{\max},\;
    \|\mathbf{x}_q(t)-\mathbf{x}_r(t)\|\!\ge\!d_{\min},
    \label{eq:dynamics}
\end{equation}
where $\mathbf{x}_r(t)$ denotes the position of any neighboring UAV $u_r$.

Because this problem is combinatorial and non-convex, exact solutions become intractable for large $N$ and $Q$.
Hence, a \textbf{GA–RF} is used to generate near-optimal expert demonstrations that serve as training data for the proposed learning-based framework \cite{Arshid2025}.
\vspace{-3mm}
%
\section{Proposed Framework}
The proposed approach transforms the deterministic MTSP formulation into a self-learning, probabilistic decision system governed by the \textit{principle of Active Inference}. It consists of two tightly coupled phases: (i) an \textbf{offline learning phase}, where a hierarchical world model is learned from expert demonstrations generated by the GA--RF optimizer, and (ii) an \textbf{online active inference phase}, where UAVs perform adaptive trajectory planning and belief updating in real time.
\vspace{-2mm}
%
\subsection{Expert Demonstrations via GA--RF}
The GA--RF jointly optimizes mission allocation, route ordering, and trajectory feasibility. Each chromosome encodes a multi-UAV solution, while repulsion forces impose penalties whenever the inter-UAV distance approaches $d_{\min}$. For each mission instance $D_m$, the optimizer yields an expert demonstration
%
%
$\tau_m=\{D_m,\mathcal{A}_m,\mathcal{P}_m,\mathcal{M}_m\}$,
where $D_m$ defines the cities and UAVs, $\mathcal{A}_m$ the mission allocation (division of cities), $\mathcal{P}_m$ the visiting order of each UAV, and $\mathcal{M}_m$ the corresponding motion trajectories.
These demonstrations form a structured dataset describing how global missions are decomposed into ordered routes and feasible motions. They serve as the empirical foundation for constructing the generative world model.
\vspace{-2mm}
%
\subsection{Hierarchical Symbolic World Model}
Each expert demonstration $\tau_m$ is transformed into a \emph{hierarchical symbolic representation} that captures the UAV swarm behavior across multiple abstraction levels. This representation organizes knowledge through three interconnected symbolic dictionaries: \emph{Mission}, \emph{Route}, and \emph{Motion}, forming the \textbf{World Model}.
\vspace{-3mm}
\paragraph*{Mission Dictionary (high level).}
Let $C=\{c_1,\dots,c_N\}$ be the set of target locations (atomic symbols or \emph{letters}). The GA--RF divides $C$ into $Q$ disjoint subsets $\mathcal{C}_1,\dots,\mathcal{C}_Q$, where each subset $\mathcal{C}_q$ represents the cities assigned to UAV $q$. The subset is encoded as a \emph{Mission Word} $\mathcal{W}^{(\text{mission})}_q=\{c\in\mathcal{C}_q\}$, and the set of all Mission Words within a mission forms a \emph{Mission Phrase}. The collection of all phrases obtained from expert data constitutes the \emph{Mission Dictionary} $\mathcal{D}_{\text{Msn}}$, modeled probabilistically as $p(\mathcal{W}^{(\text{mission})}|\mathcal{D})$.
\vspace{-3mm}
\paragraph*{Route Dictionary (mid level).}
For each Mission Word, the expert provides an ordered sequence of visits $\pi_q=[c_{q,1},\dots,c_{q,|C_q|}]$ defining the UAV's optimal traversal order. This sequence is encoded as a \emph{Route Word} $\mathcal{W}^{(\text{route})}_q$, and the collection of such words across missions defines the \emph{Route Dictionary} $\mathcal{D}_{\text{Rte}}$. This level captures the probabilistic mapping $p(\mathcal{W}^{(\text{route})}|\mathcal{W}^{(\text{mission})})$, describing how high-level task divisions are realized as ordered routes.
\vspace{-3mm}
%
\paragraph*{Motion Dictionary (low level).}
At the motion level, each consecutive pair $(c_{q,i}\!\to\!c_{q,i+1})$ generates a trajectory segment $\gamma_{q,i}(t)$, which is transformed into a feature vector $\phi(\gamma)$ describing kinematic properties such as velocity, curvature, and heading rate. Clustering these features yields a finite alphabet of \emph{Motion Letters}, and the concatenation of letters along a route forms a \emph{Motion Word} that characterizes the UAV's local dynamic behavior. Two principal motion categories are identified:
\\[2pt]
\textbf{-- Attractive Motion Letters:}
goal-directed segments dominated by the attractive potential
%
%
%
$U_{\text{att}}(\mathbf{x};\mathbf{p})
=\tfrac{1}{2}k_{\text{att}}\|\mathbf{x}-\mathbf{p}\|^2$,
which drives the UAV toward its target position $\mathbf{p}$.
\\[2pt]
\textbf{-- Repulsive Motion Letters:}
avoidance segments dominated by the repulsive potential
%
%
%
%
%
{\small
\begin{equation}
\begin{aligned}
U_{\text{rep}}(\mathbf{x}) =
\sum_{o \in \mathcal{O}} \tfrac{1}{2} k_{\text{rep}}^{(o)}
\!\left[\max\!\Big(0, \tfrac{1}{d_o(\mathbf{x})}-\tfrac{1}{d_0}\Big)\right]^2
\\
+ \sum_{\substack{r=1 \\ r \neq q}}^{Q}
\tfrac{1}{2} k_{\text{rep}}^{(\text{uav})}
\!\left[\max\!\Big(0, \tfrac{1}{d_r(\mathbf{x})}-\tfrac{1}{d_0}\Big)\right]^2,
\end{aligned}
\end{equation}
}
where $\mathcal{O}$ denotes the set of obstacles, $d_o(\mathbf{x})$ and $d_r(\mathbf{x})$ are the distances from the UAV to obstacle $o$ and to another UAV $r$, respectively, and $d_0$ is the safety distance beyond which the repulsive influence vanishes.
The coefficients $k_{\text{rep}}^{(o)}$ and $k_{\text{rep}}^{(\text{uav})}$ are the corresponding repulsion gains.

The UAV motion evolves according to the gradient flow of the total potential field:
%
%
$\dot{\mathbf{x}} = -K\nabla \big(U_{\text{att}}(\mathbf{x}) + U_{\text{rep}}(\mathbf{x})\big)$.
A trajectory segment is labeled as \emph{attractive} or \emph{repulsive} depending on the ratio of potential energies.
%
%
All Motion Words extracted from the demonstrations constitute the \emph{Motion Dictionary} $\mathcal{D}_{\text{Mot}}$, whose statistical dependence on the Route Dictionary is modeled as $p(\mathcal{W}^{(\text{motion})}|\mathcal{W}^{(\text{route})})$.
\vspace{-3mm}
\paragraph*{Hierarchical probabilistic coupling.}
The three dictionaries are linked through learned transition operators $T_{\text{Msn}\to\text{Rte}}$ and  $T_{\text{Rte}\to\text{Mot}}$.
The overall hierarchical generative model factorizes as
%
%
$p(\mathcal{W}^{(\text{motion})},
\mathcal{W}^{(\text{route})},
\mathcal{W}^{(\text{mission})}|\mathcal{D})
= \\
p(\mathcal{W}^{(\text{mission})}|\mathcal{D})\,
p(\mathcal{W}^{(\text{route})}|\mathcal{W}^{(\text{mission})})\,
p(\mathcal{W}^{(\text{motion})}|\mathcal{W}^{(\text{route})}),$
which serves as the generative prior for Active Inference during online decision-making. In addition, it encodes how global mission assignments evolve into ordered routes and, finally, into dynamically feasible motion behaviors.


%
%
\vspace{-3mm}
%
\subsection{Online Decision-Making via Active Inference}
During online operation, the UAV swarm continuously interprets sensory observations $o_t=\{C_t,\mathbf{X}_t\}$, where $C_t$ denotes the set of cities (letters) and $\mathbf{X}_t=\{\mathbf{x}_c\}_{c\in C_t}$ their coordinates. The swarm relies on the World Model to infer the most plausible symbolic configuration $s_t=\{\mathcal{W}^{(\text{mission})}_t, \mathcal{W}^{(\text{route})}_t, \mathcal{W}^{(\text{motion})}_t\}$ and to decide subsequent actions that preserve coherence with prior knowledge while adapting to new environmental realizations. Decision-making is performed hierarchically across three action levels: mission division, route ordering, and motion generation.

The World Model provides probabilistic reference distributions learned from expert demonstrations:
$ p_{\text{ref}}^{(\text{Msn})} = p(\mathcal{W}^{(\text{mission})}|\mathcal{D}),
\quad p_{\text{ref}}^{(\text{Rte})} = p(\mathcal{W}^{(\text{route})}|\mathcal{W}^{(\text{mission})}),
\quad p_{\text{ref}}^{(\text{Mot})} = p(\mathcal{W}^{(\text{motion})}|\mathcal{W}^{(\text{route})}).$

At each level $\ell\!\in\!\{\text{Msn},\text{Rte},\text{Mot}\}$, the UAV evaluates candidate actions $a^{(\ell)}$ and selects the one minimizing the divergence between the predicted posterior and the corresponding reference distribution:
\begin{equation}
\small
\mathcal{A}_{\ell}(a^{(\ell)}) =
D_{\text{KL}}\!\Big(
q_t(\mathcal{W}^{(\ell)}|a^{(\ell)}) \,\big\|\,
p_{\text{ref}}^{(\ell)}
\Big).
\label{eq:abn_general}
\end{equation}
This hierarchical abnormality minimization ensures that decisions remain consistent with the statistical structure encoded in the world model.

\paragraph*{(1) Division-level decision (Mission Words).}
At the highest level, the prior $p_{\text{ref}}^{(\text{Msn})}$, derived from the transition matrix $T_{D\rightarrow\text{Msn}}$, provides \emph{reference Mission Words} representing typical partitions of cities among UAVs. A division action $a^{(\text{Msn})}$ specifies a partition $\mathcal{W}^{(\text{mission})}=\{\mathcal{C}_1,\dots,\mathcal{C}_Q\}$, and the selected action is
\begin{equation}
\scriptsize
a^{(\text{Msn})\!*}
= \arg\min_{a^{(\text{Msn})}}
D_{\text{KL}}\!\Big(
q_t(\mathcal{W}^{(\text{mission})}|a^{(\text{Msn})})
\;\big\|\;
p_{\text{ref}}^{(\text{Msn})}\Big).
\label{eq:msn_sel}
\end{equation}
If a new city $c^\star$ appears, the swarm determines which UAV should serve it by minimizing its divergence from the reference Mission Words:
\begin{equation}
\scriptsize
q^\star = \arg\min_{q\in\{1,\dots,Q\}}
D_{\text{KL}}\!\Big(
q_t(\mathcal{W}^{(\text{mission})}\cup\{c^\star\}\!\in\!\mathcal{C}_q)
\;\big\|\;
p_{\text{ref}}^{(\text{Msn})}\Big).
\label{eq:new_city_assignment}
\end{equation}
Equation \eqref{eq:new_city_assignment} ensures that new cities are assigned to the subset producing the smallest deviation from the reference Mission Words.
\vspace{-3mm}
%
\paragraph*{(2) Ordering-level decision (Route Words).}
Given the selected division, the prior $p_{\text{ref}}^{(\text{Rte})}$---from $T_{\text{Msn}\rightarrow\text{Rte}}$-- provides \emph{reference Route Words} describing the expected visiting order of each UAV. For UAV $q$ with subset $\mathcal{C}_q$, a candidate ordering action $a^{(\text{Rte})}$ defines a permutation
$\mathcal{W}^{(\text{route})}_q
=[c_{q,1},\dots,c_{q,|\mathcal{C}_q|}]$,
and the chosen order minimizes
\begin{equation}
\small
a^{(\text{Rte})\!*}
= \arg\min_{a^{(\text{Rte})}}
D_{\text{KL}}\!\Big(
q_t(\mathcal{W}^{(\text{route})}|a^{(\text{Rte})})
\;\big\|\;
p_{\text{ref}}^{(\text{Rte})}\Big).
\label{eq:rte_sel}
\end{equation}
When a new city $c^\star$ is introduced within a route, it is first assigned to its subset by \eqref{eq:new_city_assignment}, and its insertion position $j^\star$ is then determined by scanning all possible positions to find
\begin{equation}
\small
j^\star = \arg\min_{j}
D_{\text{KL}}\!\Big(
q_t(\mathcal{W}^{(\text{route})}\cup\{c^\star\}\text{ at }j)
\;\big\|\;
p_{\text{ref}}^{(\text{Rte})}\Big).
\label{eq:insert_pos}
\end{equation}
Equations \eqref{eq:new_city_assignment} and \eqref{eq:insert_pos} separately govern subset assignment and position selection, maintaining consistency with the reference Route Words.

\paragraph*{(3) Motion-level decision (Motion Words).}
At the lowest level, the prior $p_{\text{ref}}^{(\text{Mot})}$---derived from $T_{\text{Rte}\rightarrow\text{Mot}}$---provides \emph{reference Motion Words} composed of symbolic motion letters that describe short-term behaviors such as attractive or repulsive flight segments. Each motion action $a^{(\text{Mot})}$ corresponds to a candidate motion word selected from the Motion Dictionary $\mathcal{D}_{\text{Mot}}$. The selected motion policy minimizes
\begin{equation}
\small
a^{(\text{Mot})\!*}
= \arg\min_{a^{(\text{Mot})}}
D_{\text{KL}}\!\Big(
q_t(\mathcal{W}^{(\text{motion})}|a^{(\text{Mot})})
\;\big\|\;
p_{\text{ref}}^{(\text{Mot})}\Big),
\label{eq:mot_sel}
\end{equation}
ensuring that instantaneous dynamics remain coherent with the reference Motion Words. When the state estimator (e.g., EKF) predicts possible collisions or obstacles, abnormality in \eqref{eq:mot_sel} increases, prompting a policy switch toward motion words with higher likelihood of repulsive motion letters.

\paragraph*{Hierarchical integration.}
The three action levels are coupled through the World Model factorization:
\begin{equation}
p(\mathcal{W}^{(\text{motion})},
\mathcal{W}^{(\text{route})},
\mathcal{W}^{(\text{mission})}|\mathcal{D})
=
p_{\text{ref}}^{(\text{Msn})}\,
p_{\text{ref}}^{(\text{Rte})}\,
p_{\text{ref}}^{(\text{Mot})}.
\end{equation}
The global objective integrates the level-wise divergence measures:
\begin{equation}
\scriptsize
\mathcal{A}_{\text{total}}
=
\mathcal{A}_{\text{Msn}}
+
\mathcal{A}_{\text{Rte}}
+
\mathcal{A}_{\text{Mot}},
\mathcal{A}_{\ell}
= D_{\text{KL}}\!\big(
q_t(\mathcal{W}^{(\ell)}|a^{(\ell)})
\big\|
p_{\text{ref}}^{(\ell)}
\big).
\label{eq:total_abn}
\end{equation}
Minimizing \eqref{eq:total_abn} yields a globally consistent inference process across all abstraction levels, enabling the swarm to reorganize mission assignments, update route orders, and refine motion behaviors in response to novel environmental realizations while maintaining coherence with the reference Mission, Route, and Motion Words encoded in the world model.

\vspace{-3mm}
%
\section{Results}
We evaluate the proposed Active Inference framework in an online multi-UAV trajectory design setting. Multiple UAVs operate in a \(1000\times1000\)~m area at a fixed altitude of 200~m. For training, \(M=5000\) missions with 50 randomly placed targets are generated via GA–RF, yielding optimal velocity and visiting-order sequences that are symbolically encoded to build the world model. In testing, unseen missions are solved using only the learned model. A modified Q-Learning (QL) method trained on the same data is used as a baseline.

\begin{figure}
    \begin{minipage}[b]{0.48\linewidth}
     \centering
        \includegraphics[width=4.5cm]{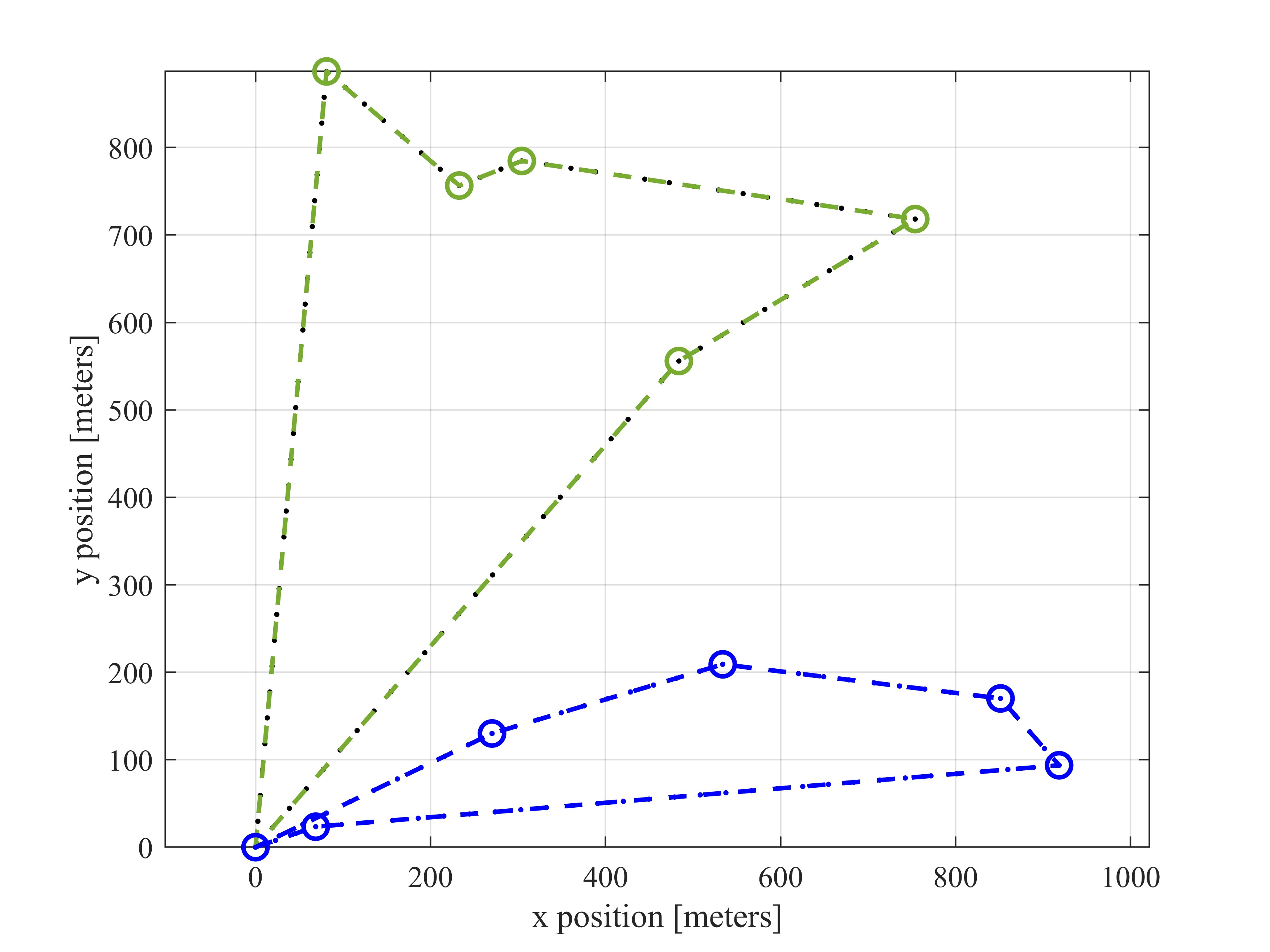}
        \centerline{\scriptsize (a)}
    \end{minipage}
    \begin{minipage}[b]{0.48\linewidth}
     \centering
        \includegraphics[width=4.5cm]{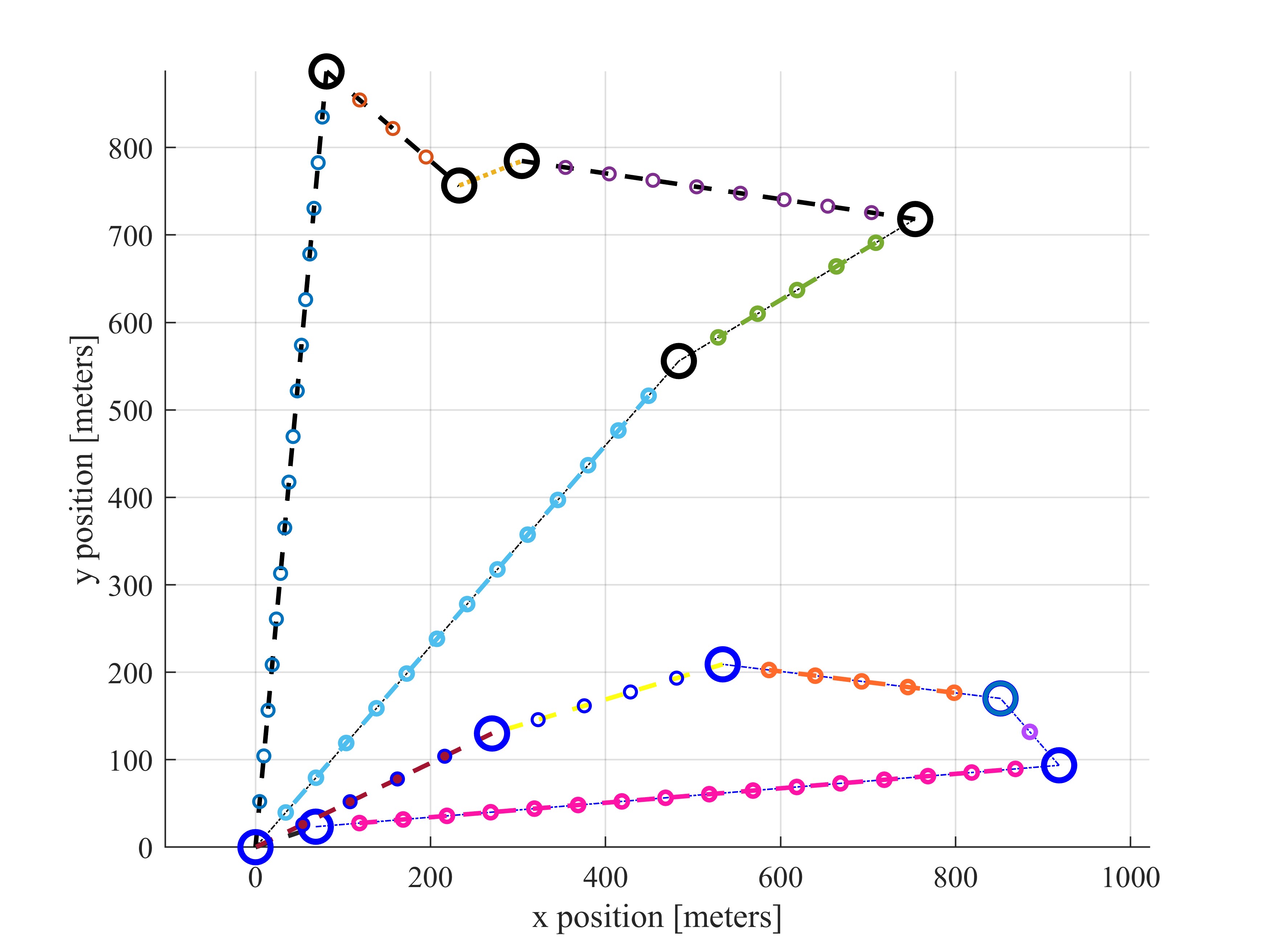}
        \centerline{\scriptsize (b)}
    \end{minipage}
    \caption{Hierarchical behavior: (a) mission division and routing; (b) local motion via learned motion words.}
    \label{Fig_example_AIn_levels}
\end{figure}

\begin{figure}
    \begin{minipage}[b]{0.48\linewidth}
     \centering
        \includegraphics[width=4.5cm]{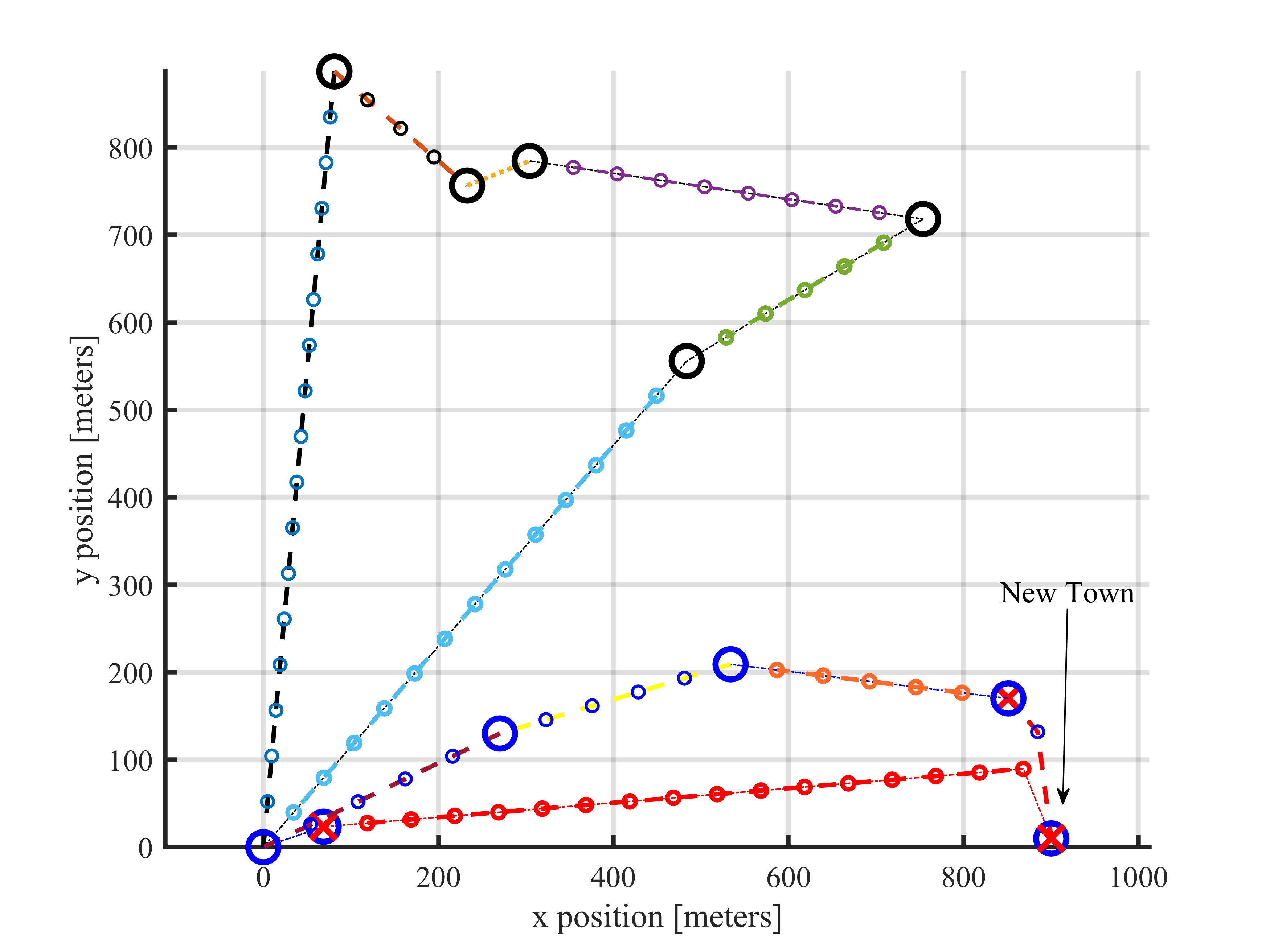}
        \centerline{\scriptsize (a) }
    \end{minipage}
    \begin{minipage}[b]{0.48\linewidth}
     \centering
        \includegraphics[width=4.5cm]{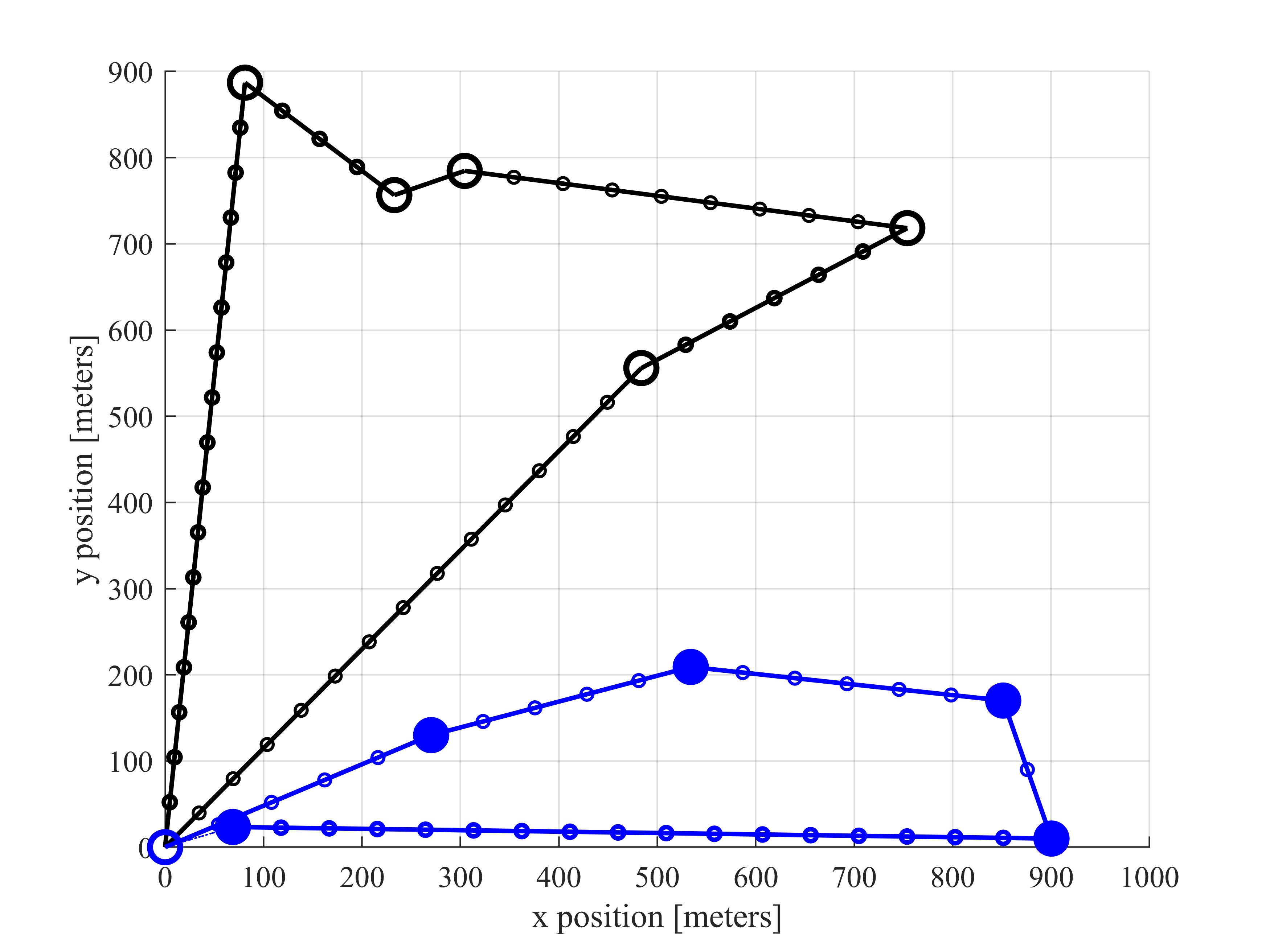}
        \centerline{\scriptsize (b) }
    \end{minipage}
    \caption{Adaptive replanning to a new target: (a) surprise; (b) belief update and corrected path.}
    \label{Fig_example_surprisingSituation}
\end{figure}

\begin{figure}[ht!]
    \begin{minipage}[b]{0.48\linewidth}
     \centering
        \includegraphics[width=4.5cm]{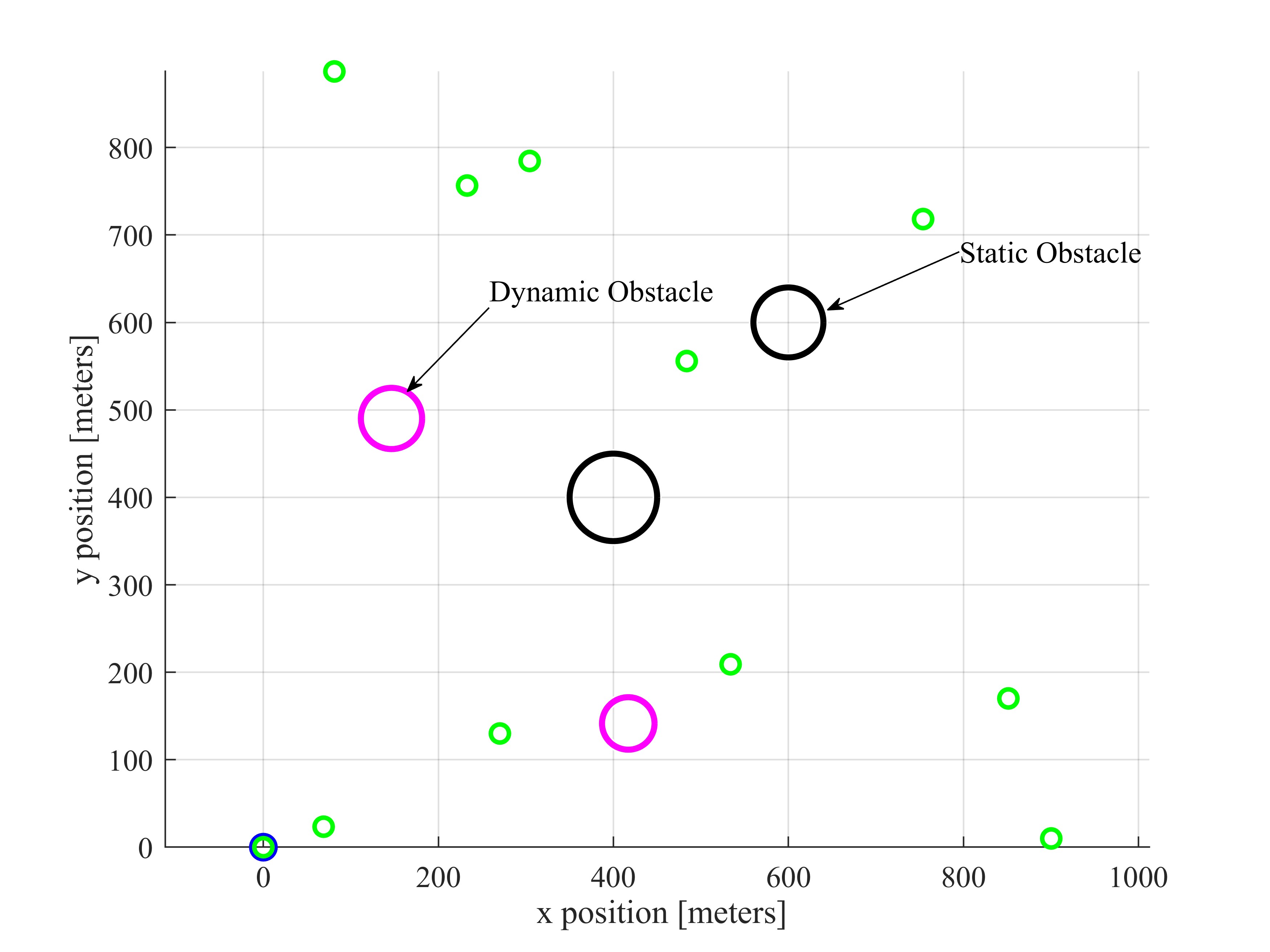}
        \centerline{\scriptsize (a)}
    \end{minipage}
    \begin{minipage}[b]{0.48\linewidth}
     \centering
        \includegraphics[width=4.5cm]{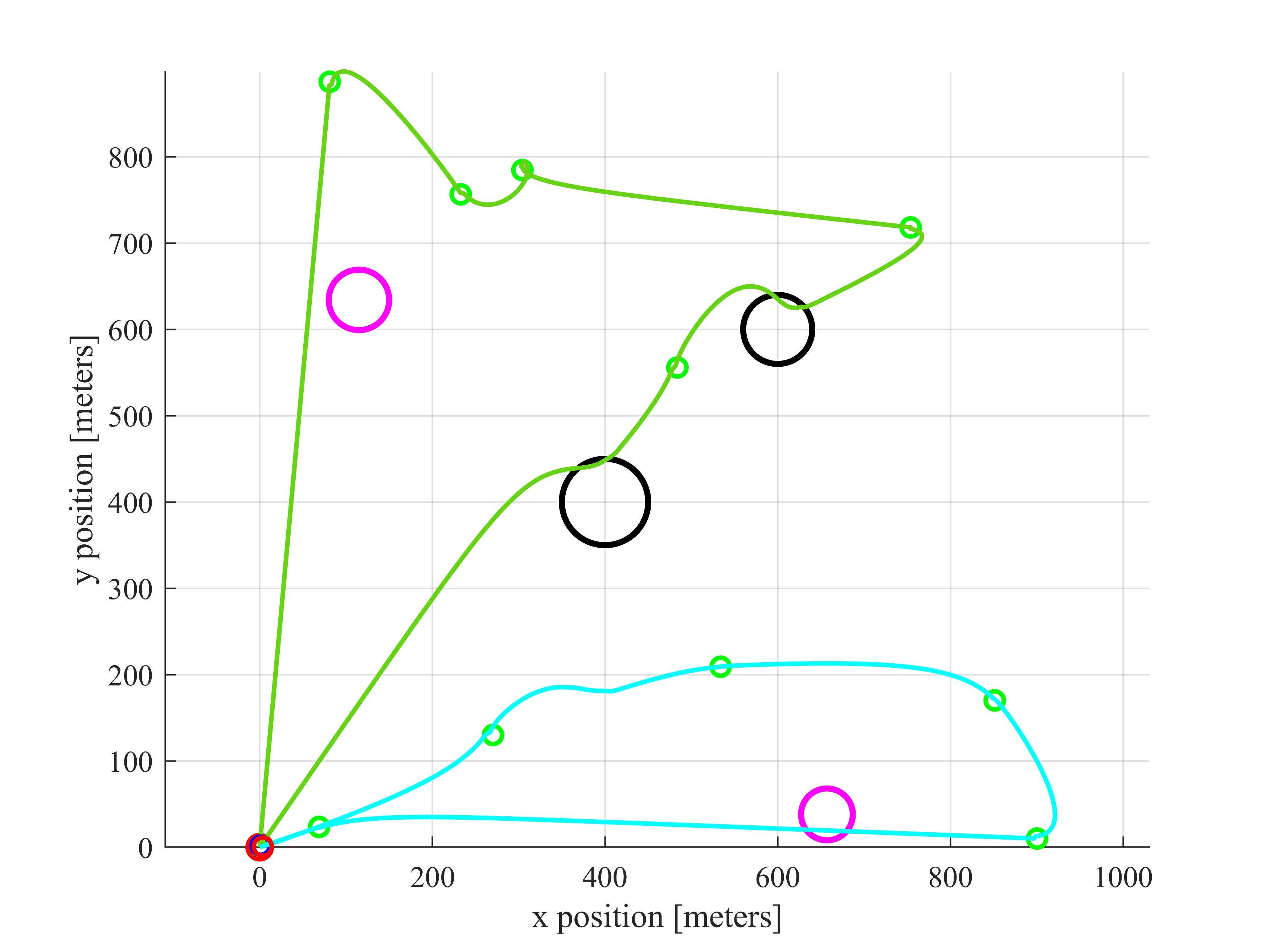}
        \centerline{\scriptsize (b)}
    \end{minipage}
    \caption{EKF-assisted correction and obstacle avoidance: (a) prediction/correction; (b) static (black) and dynamic (pink) obstacles.}
    \label{Fig_example_obstacles}
\end{figure}

\begin{figure}[ht!]
    \begin{minipage}[b]{0.48\linewidth}
     \centering
        \includegraphics[width=4.5cm]{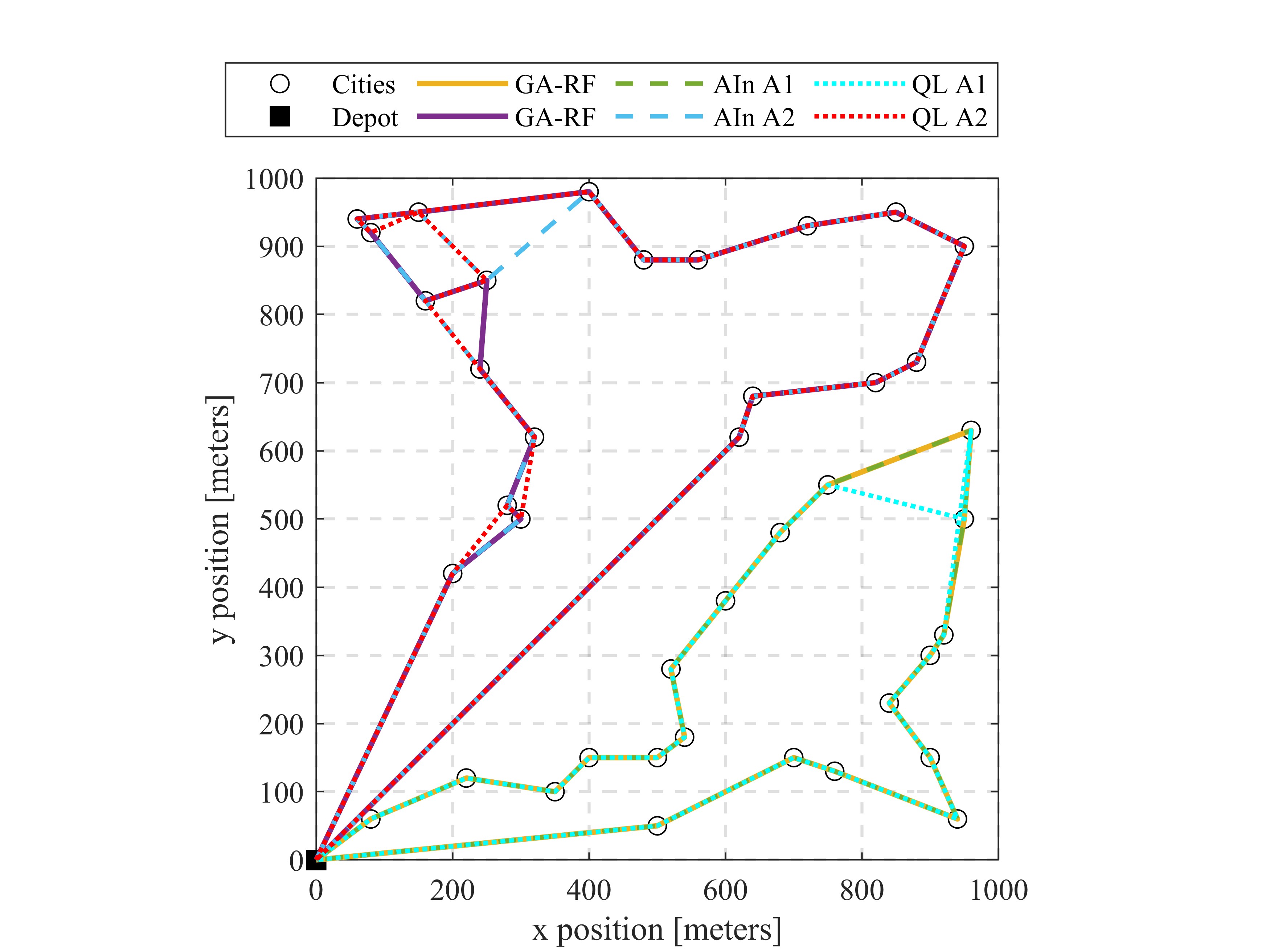}
        \centerline{\scriptsize (a) 40 Towns }
    \end{minipage}
    \begin{minipage}[b]{0.48\linewidth}
     \centering
        \includegraphics[width=4.5cm]{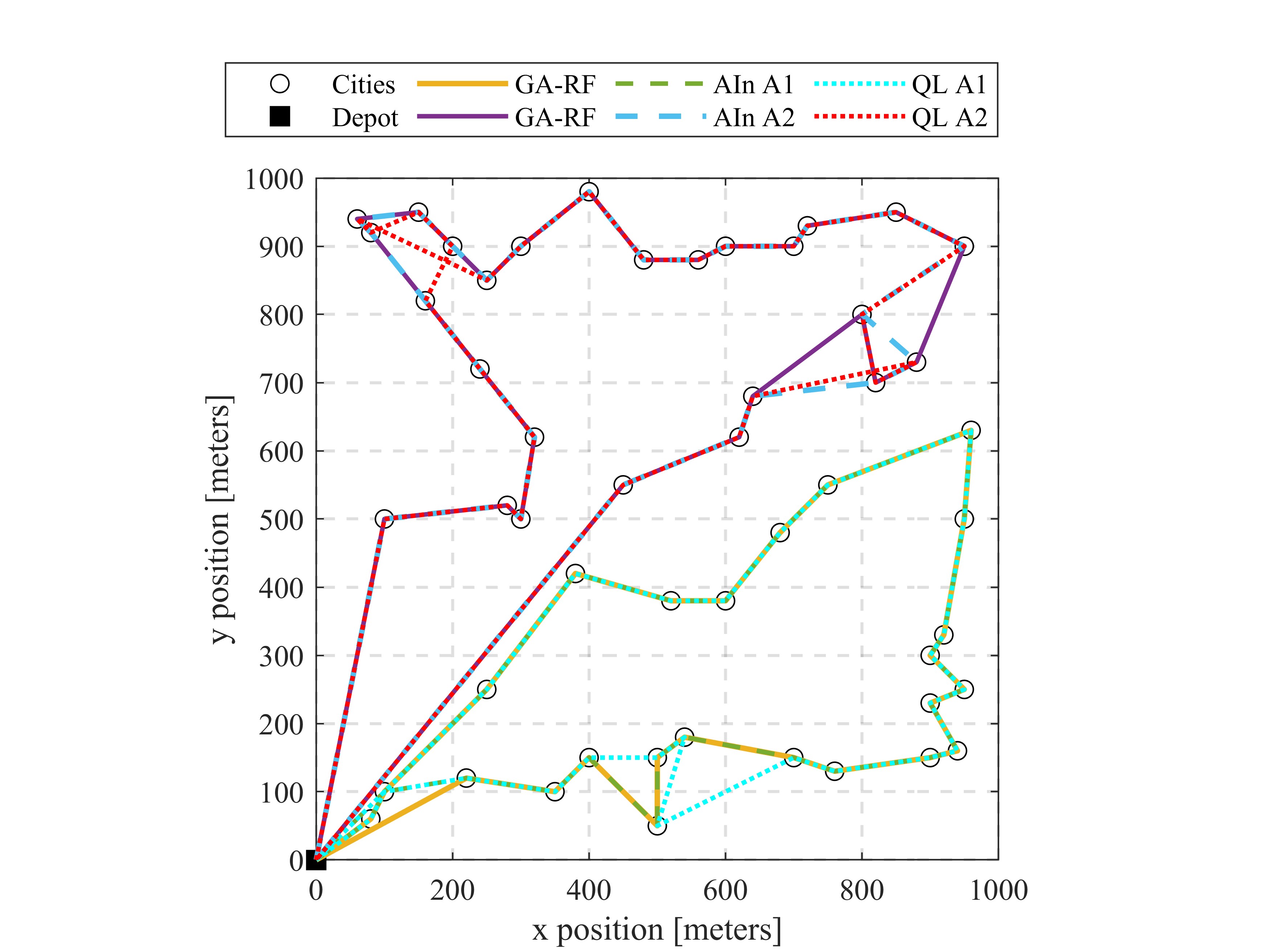}
        \centerline{\scriptsize (b) 50 Towns}
    \end{minipage}
    \caption{Trajectories: Active Inference vs. Modified QL.}
    \label{Fig_example_comparisonTraj}
\end{figure}

\begin{figure}[ht!]
    \begin{minipage}[b]{0.48\linewidth}
     \centering
        \includegraphics[width=4.5cm]{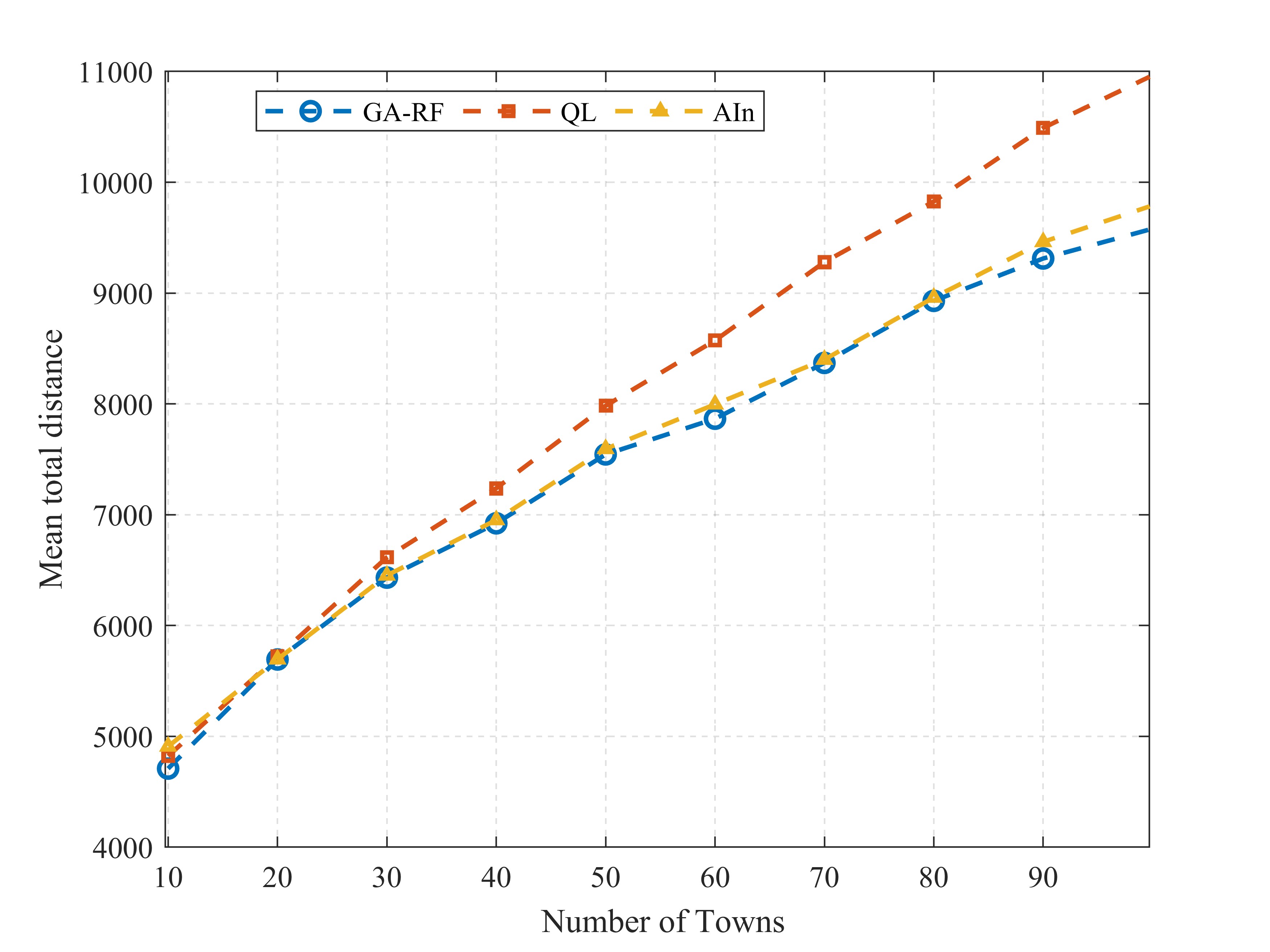}
        \centerline{\scriptsize (a) Completion time}
    \end{minipage}
    \begin{minipage}[b]{0.48\linewidth}
     \centering
        \includegraphics[width=4.5cm]{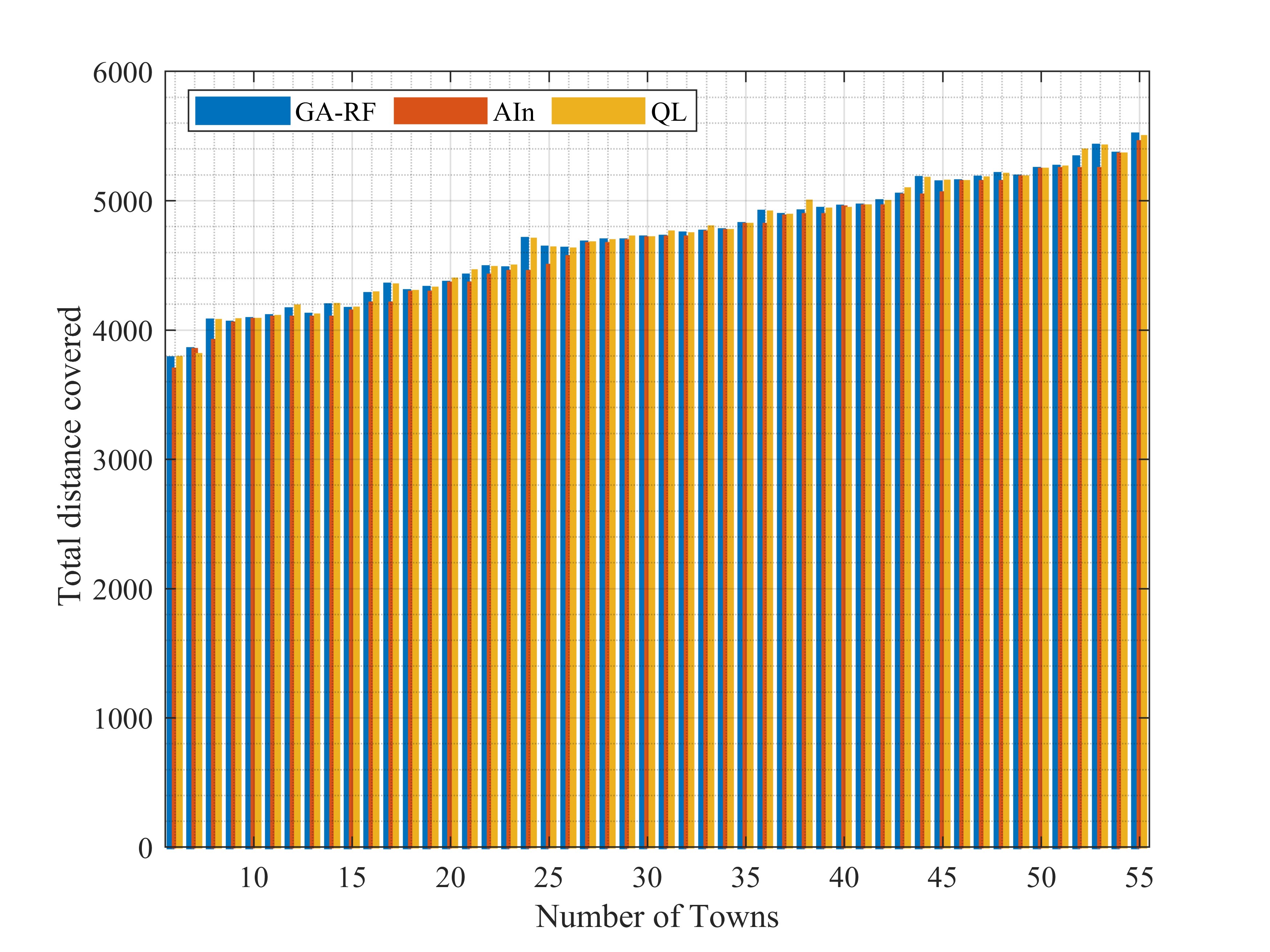}
        \centerline{\scriptsize (b) Covered distance}
    \end{minipage}
    \caption{Quantitative comparison: (a) time; (b) distance.}
    \label{Fig_comparison_completionTime_Distance}
\end{figure}

Fig.~\ref{Fig_example_AIn_levels} illustrates the hierarchical behavior: high/mid levels perform mission allocation and route ordering consistent with Mission/Route Words, while the motion level executes attractive/repulsive segments that realize dynamically feasible paths.
When a new target appears, prediction–action mismatch triggers belief revision and route re-optimization. As shown in Fig.~\ref{Fig_example_surprisingSituation}, the UAV adapts by minimizing divergence from the world model, restoring mission consistency in real time.
To improve smoothness and safety, an EKF is integrated for local state estimation and obstacle avoidance. Fig.~\ref{Fig_example_obstacles} shows EKF-based prediction/correction and collision-free navigation.
Fig.~\ref{Fig_example_comparisonTraj} compares trajectories: Active Inference maintains belief–action consistency and smoother paths, whereas QL deviates at the motion level. Quantitative results in Fig.~\ref{Fig_comparison_completionTime_Distance} show lower mission time and total distance than both GA–RF and QL, confirming improved efficiency and generalization.

Overall, the framework achieves faster convergence, higher stability, and stronger adaptability than QL, validating autonomous probabilistic reasoning and real-time trajectory re-planning in complex environments.
\vspace{-3mm}
\section{Conclusion}
This paper presented an Active Inference–based framework for adaptive UAV swarm trajectory design. The method learns a probabilistic world model from expert demonstrations and employs it for hierarchical decision-making across mission, route, and motion levels. Continuous belief updating enables adaptive and energy-efficient coordination, while an EKF-assisted module ensures smooth and collision-free navigation. Simulations show faster convergence and higher stability than modified QL, establishing a unified probabilistic basis for scalable and cognitively adaptive UAV swarm control.

\vfill\pagebreak

\bibliographystyle{IEEEbib}
\bibliography{strings,refs}

\end{document}